\PassOptionsToPackage{dvipsnames,table}{xcolor}
\documentclass{bmvc2k}
\usepackage[T1]{fontenc}
\usepackage{microtype}

\usepackage{etoolbox}

\usepackage{booktabs}
\usepackage{tabularray}
\usepackage{adjustbox}
\UseTblrLibrary{
    booktabs,
    siunitx,
}

\SetTblrInner{
  columns = {colsep=4pt},
  rows= {rowsep=.25pt}
}

\usepackage{amsfonts}
\usepackage{amsmath}
\usepackage{amssymb}
\usepackage{scalerel}
\usepackage{nicefrac}
\usepackage{bm}

\usepackage{graphicx}
\usepackage{placeins}

\DeclareMathOperator*{\argmin}{arg\,min}
\newcommand*{\plh}{{\mkern 2mu\bm{\cdot}\mkern 2mu}}
\DeclareMathOperator*{\E}{\mathbb{E}}

\let\oldtimes\times
\let\singletimes\oldtimes
\def\times{{\mkern1mu\oldtimes\mkern1mu}}% small space

\usepackage{xcolor}

\usepackage{caption}
\usepackage{tikz}
\usetikzlibrary{
  arrows,
  arrows.meta,
  backgrounds,
  calc,
  fit,
  positioning,
}
\usepackage{pgfplots}
\usepackage{pgfplotstable}

\usepgfplotslibrary{
  colorbrewer,
  colormaps,
  fillbetween,
  groupplots,
}
\pgfplotsset{compat=1.18,
  my footnotesize/.style={
    footnotesize,
    legend style={font=\footnotesize},
    tick label style={font=\scriptsize},
    label style={font=\footnotesize},
    title style={font=\footnotesize},
  },
}

\pgfdeclarelayer{background}
\pgfdeclarelayer{foreground}
\pgfsetlayers{background,main,foreground}

\makeatletter
\tikzset{
  /addpiecewise/.cd,
  color/.store in={\@color},
  path options/.store in={\@pathoptions},
  markers options/.store in={\@markersoptions},
  /tikz/.cd,
}
\newcommand{\addpiecewise}[2][]{
  \tikzset{
    /addpiecewise/.cd,
    path options={}, % init
    markers options={}, % init
    color=black, % init
    #1,% read the arguments
    /tikz/.cd}
  \def\firstitem{1}
  \xdef\markercoords{}
  
  \foreach \px/\py in {#2} {
    \xdef\markercoords{\markercoords (\px, \py, {f(\px,\py)})}
    
    \ifnum\firstitem=1
    \xdef\lastx{\px}
    \xdef\lasty{\py}
    \xdef\firstitem{0}
    \else
    \edef\temp{
      \noexpand\addplot3 [
      variable=t,
      domain=0:1,
      samples=25, 
      samples y=0,% forcing it 1d
      very thick,
      draw opacity=0.85,
      \@pathoptions,
      \@color!95!black,
      ] (
      {\lastx + t*(\px - \lastx)},
      {\lasty + t*(\py - \lasty)},
      {f( (\lastx + t*(\px - \lastx)), (\lasty + t*(\py - \lasty)) )}
      );
    }
    \temp
    
    \xdef\lastx{\px}
    \xdef\lasty{\py}
    \fi
  }
  
  \edef\drawmarkers{
    \noexpand\addplot3 [
    only marks,
    mark options={scale=0.9, fill opacity=0.75, draw opacity=1, draw=\@color!75!black},
    mark=*,
    \@color,
    \@markersoptions
    ] coordinates {\markercoords};
  }
  \drawmarkers
}
\makeatother

\pgfplotsset{
  filled plot/.style={
      mark size=1pt,
      fill opacity=0.5,
  },
}
\colorlet{vit}{Dark2-D}
\colorlet{brh}{Dark2-A}
\colorlet{ours}{Dark2-B}
\colorlet{toast}{Dark2-C}

\usepackage[inline]{enumitem}

\usepackage{etoolbox}
\makeatletter
\patchcmd{\NAT@citexnum}
  {\@citea \NAT@test{1}\ \NAT@mbox{\NAT@@open}}
  {\@citea \NAT@test{1}~\NAT@mbox{\NAT@@open}}
  {}{}
\patchcmd{\NAT@citex}
  {\hyper@natlinkbreak{\ \NAT@@open\if*#1*\else#1\ \fi}}
  {\hyper@natlinkbreak{~\NAT@@open\if*#1*\else#1\ \fi}}
  {}{}
\makeatother

\makeatletter
\patchcmd{\NAT@test}{\else \NAT@nm}{\else \NAT@nmfmt{\NAT@nm}}{}{}

\DeclareRobustCommand\citepos
  {\begingroup
   \let\NAT@nmfmt\NAT@posfmt% ...except with a different name format
   \NAT@swafalse\let\NAT@ctype\z@\NAT@partrue
   \@ifstar{\NAT@fulltrue\NAT@citetp}{\NAT@fullfalse\NAT@citetp}}

\let\NAT@orig@nmfmt\NAT@nmfmt
\def\NAT@posfmt#1{\NAT@orig@nmfmt{#1's}}
\makeatother

\makeatletter
\def\NAT@spacechar{~}% NEW
\makeatother

\usepackage{hyperref}
\usepackage{url}

\hypersetup{
  pdfborder={0 0 0},
  breaklinks,
  colorlinks,
  linkcolor = BrickRed,
  citecolor = RoyalBlue,
  urlcolor  = WildStrawberry,
}

\usepackage{cleveref}
\crefname{figure}{Fig.}{Figs.}
\Crefname{figure}{Fig.}{Figs.}
\crefformat{equation}{Eq.~(#2#1#3)}
\Crefformat{equation}{Eq.~(#2#1#3)}

\colorlet{highlight}{orange!10}

\usepackage{xspace}
\makeatletter
\DeclareRobustCommand\onedot{\futurelet\@let@token\@onedot}
\def\@onedot{\ifx\@let@token.\else.\null\fi\xspace}

\def\ie{{i.e}\onedot} 
\def\cf{{cf}\onedot}

\makeatother

\usepackage{datatool}

\DTLloaddb{natimg}{data/natvals.txt}
\newcommand{\NatVal}[2]{\DTLfetch{natimg}{id}{#1}{#2}}
\newcommand{\NatDelta}[2]{\DTLfetch{natimg}{id}{#1}{#2_delta}}

\makeatletter
\newcommand{\equalcontributionnote}{%
    \patchcmd{\maketitle}
    {\hrule\vskip\baselineskip}
    {\hrule\vskip\baselineskip\BMVA@blfootnote{\hspace{-1.9em}\textsuperscript{*}Equal contribution.}}
    {\message{^^J--> Patch bmv@maketitle successful!^^J}}
    {\message{^^J--> Patch bmv@maketitle failed!^^J}}
}
\newcommand{\githubnote}[1]{%
    \patchcmd{\maketitle}
    {\hrule\vskip\baselineskip}
    {\hrule\vskip\baselineskip\BMVA@blfootnote{\hspace{-1.9em}Code available at: {\scriptsize\url{#1}}.}}% for some reason \href doesn't work (it is not linking the url), dunno why :( 
    {\message{^^J--> Patch bmv@maketitle successful!^^J}}
    {\message{^^J--> Patch bmv@maketitle failed!^^J}}
}
\makeatother

\makeatletter
\newcommand{\method}{Transformer-Within-Transformer\xspace}
\newcommand{\shortmethod}{TWT\@ifnextchar.{\@}{\xspace}}
\makeatother
\robustify{\shortmethod}

\newcommand{\appref}[1]{Appendix~\ref{#1}}

\title{A Smaller Transformer in Your Transformer}

\addauthor{Dhananjay Tomar}{dhananjt@ifi.uio.no}{1,2,*}
\addauthor{Marius Aasan}{mariuaas@uio.no}{2,3,*}
\addauthor{Andreas Kleppe}{andrekle@ifi.uio.no}{1,2,3}
\addauthor{Adín Ramírez Rivera}{adinr@uio.no}{2,3}

\addinstitution{
Institute for Cancer Genetics and Informatics\\
Oslo University Hospital\\
Oslo, Norway
}

\addinstitution{
Department of Informatics\\
University of Oslo\\
Oslo, Norway
}

\addinstitution{
SFI Visual Intelligence\\
UiT The Arctic University of Norway\\
Tromsø, Norway
}

\runninghead{Tomar \bmvaEtAl}{A Smaller Transformer in Your Transformer}

\begin{document}
\begin{filecontents*}{data/natvals.txt}
id,in_val,in_val_delta,in_real,in_real_delta,in_v2,in_v2_delta
deitv3_base,    83.1,     ,87.7,     ,71.9,
deitv3_nose,    81.5,-1.6 ,86.2,-1.5 ,70.4,-1.5
deitv3_wd_6,    82.7,-0.4 ,87.1,-0.6 ,71.1,-0.8
deitv3_wd_5,    82.1,-1.0 ,86.8,-0.9 ,70.8,-1.1
deitv3_raptor_6,82.9,-0.2 ,87.3,-0.4 ,71.5,-0.4
deitv3_raptor_5,82.2,-0.9 ,86.9,-0.8 ,70.9,-1.0
deitv3_twt_6,   82.8,-0.3 ,87.5,-0.2 ,71.4,-0.5
deitv3_twt_5,   82.1,-1.0 ,86.9,-0.8 ,71.0,-0.9
dinov2_base,    84.6,     ,88.5,     ,74.9,
dinov2_nose,    82.8,-1.8 ,87.0,-1.5 ,72.1,-2.8
dinov2_wd_6,    83.5,-1.1 ,87.5,-1.0 ,73.6,-1.3
dinov2_wd_5,    82.4,-2.2 ,86.4,-2.1 ,73.4,-1.5
dinov2_raptor_6,84.0,-0.6 ,87.8,-0.7 ,74.6,-0.3
dinov2_raptor_5,83.4,-1.2 ,87.2,-1.3 ,74.0,-0.9
dinov2_twt_6,   84.2,-0.4 ,88.1,-0.4 ,74.6,-0.3
dinov2_twt_5,   82.4,-2.2 ,86.9,-1.6 ,73.8,-1.1
dinov3_base,    85.2,     ,89.3,     ,75.0,
dinov3_nose,    83.2,-2.0 ,87.5,-1.8 ,73.2,-1.8
dinov3_wd_6,    84.1,-1.1 ,88.4,-0.9 ,74.3,-0.7
dinov3_wd_5,    83.3,-1.9 ,87.6,-1.7 ,73.6,-1.4
dinov3_raptor_6,84.6,-0.6 ,88.6,-0.7 ,74.6,-0.4
dinov3_raptor_5,83.7,-1.5 ,88.1,-1.2 ,74.0,-1.0
dinov3_twt_6,   84.4,-0.8 ,88.6,-0.7 ,74.3,-0.7
dinov3_twt_5,   83.4,-1.8 ,88.0,-1.3 ,73.9,-1.1
\end{filecontents*}

% Globally suppress harmless nullfont warnings caused by pgfplots fillbetween leaking during shipout
\tracinglostchars=0

\equalcontributionnote
\githubnote{https://github.com/dsb-ifi/TWT}
\maketitle

\begin{abstract}
Recent findings indicate that Vision Transformers settle into locally similar computational phases, implying a level of depthwise computational redundancy. However, existing methods to exploit this redundancy either fail to reduce inference compute or severely degrade model expressivity. In this work, we formalise a unified view of block redundancy that decouples the geometry from specific surrogate interventions. We then introduce Transformer-Within-Transformer (TWT), a post-hoc method that fuses contiguous groups of redundant layers into a single learned surrogate layer. TWT reduces parameter count and inference compute while remaining competitive with original models using half the depth on natural images, and in several downstream histopathology settings, TWT matches or even improves on the original baseline.
\end{abstract}

% -------------------------------------------------------------------------
\section{Introduction}
\label{sec:intro}

Vision Transformers (ViTs)~\citep{vaswani2017attention,oquab2024dinov2, simeoni2025dinov3} have a particular architectural austerity; after the initial token embedding, each layer acts on the same token space, updating the representation through residual attention and feed-forward networks. 
Beyond its elegance~\citep{schwarzschild2022uncanny}, this architectural homogeneity exposes a central mode of investigation: how do layers organise to perform distinct computational tasks? % Still a bit up in the air, teleological implication?

Recent discoveries~\citep{cannistraci2026toast,jacobs2026block} show that ViTs tend to settle into depthwise-contiguous computational phases with a high degree of inter-layer similarity, implying a form of \textit{computational redundancy}.
While concurrent research agrees on the identification of the symptom, the methods used to exploit local depthwise redundancy vary from local recurrence~\citep{jacobs2026block}, which lowers parameter counts but maintains the overall compute, to non-mixing approximations~\citep{cannistraci2026toast,lin2024mlp}---which reduce compute but incur a more prominent drop in performance.

Our work focuses on providing a more unified view of the phenomenon of \textit{block redundancy} in ViTs and proposes a method that better exploits redundancy with minimal performance loss.
Our contribution is threefold.
\begin{enumerate*}[label={(\roman*)}]
    \item We provide a formalisation of block redundancy in ViTs where recurrence and linear approximation appear as two sides of the same coin: both approximate local phases of computation whose constituent layers exhibit high functional similarity.
    \item We apply this paradigm with \textit{Transformer-Within-Transformer} (\shortmethod), a post-hoc method that fuses redundant blocks into a single surrogate layer, preserving the core ViT architecture.
    \item We show that \shortmethod yields parameter reductions comparable to recurrent surrogates~\cite{jacobs2026block} while reducing executed inference computation, without incurring the performance degradation of simple linear surrogates. 
    Notably, in histopathology foundation models, \shortmethod improves downstream performance relative to the original model, and can even outperform larger baselines.
\end{enumerate*}

\begin{figure}[tb]
  \centering
  \includegraphics[width=0.89\linewidth]{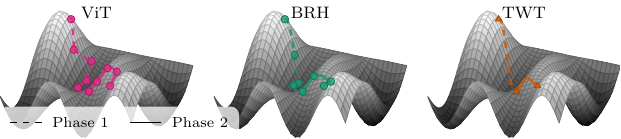}
  % \vspace{10pt}
  \caption{
   Schematic evolution of block approximations in representation space.
   Dashed and solid paths indicate two contiguous computational phases.
   (Left) A standard \textcolor{vit}{ViT} follows jagged, uneven steps through the phases encoded by its successive blocks.
   (Middle) The \textcolor{brh}{block-recurrent} approximation replaces each phase by repeated applications of a shared block, producing smoother, regular steps.
   (Right) Our \textcolor{ours}{\shortmethod} fuses each phase into a single learned surrogate transformation, bypassing the intermediate block trajectory while preserving its endpoint.
  }
  \label{fig:overview}
\end{figure}

\subsection{Note on Terminology and Nomenclature}
\label{sec:terminology}

The term \textit{block} is used somewhat indiscriminately, denoting both the elementary computational unit of a ViT and, in discussions of depthwise redundancy, a contiguous group of such units.
This quickly devolves into confusing references to \textit{blocks of blocks}.
To avoid ontological gymnastics, we elect to call the elementary ViT unit a \textit{layer}: a multi-head attention operator and a feed-forward network.
We reserve \textit{block} to mean a contiguous sequence of layers, usually grouped by a homogeneity criterion, such as cosine similarity.

\section{Block Redundancy in Vision Transformers}
\label{sec:pch}

To formalise the procedural representational flow in ViTs, we distinguish between theoretical \textit{phases} of computation and the empirical \textit{blocks} of layers learned by a network.
A \textit{phase} is an ideal, theoretical stage of computation that executes a specific set of operations on the input.
In practice, deep neural networks learn an approximation of these phases by distributing the computation across several learned layers.
Evidence of these underlying phases emerges as \textit{block patterns} in the network---contiguous groups of layers whose outputs exhibit high functional similarity.
Previous work~\citep{kornblith2019cka,cannistraci2026toast} investigates the representational flow through these blocks, observing that updates tend to periodically decelerate, and some claim this implies an underlying recurrent program~\citep{jacobs2026block}.
Our objective is to understand this representational flow by characterising and exploiting inherent redundancies in empirical block structures that capture distinct computational phases of existing models.

We posit that the contiguous blocks observed in ViTs are geometrically redundant pathways navigating a noisy manifold, and their iterative behaviour is an artefact of the optimisation landscape rather than a strict computational necessity (\cf \Cref{fig:overview}).
Consequently, these multi-step iterative blocks can be approximated by surrogate models that reflect these ideal phases. 
Our claim is that \textit{the particular choice of surrogate layer} is the distinguishing factor between different approaches, and is the target of our formalisation of block redundancy.

\begin{figure}[tb]
    \centering
    \includegraphics[
        width=\linewidth,
        trim=0 7bp 0 8bp,
        clip
    ]{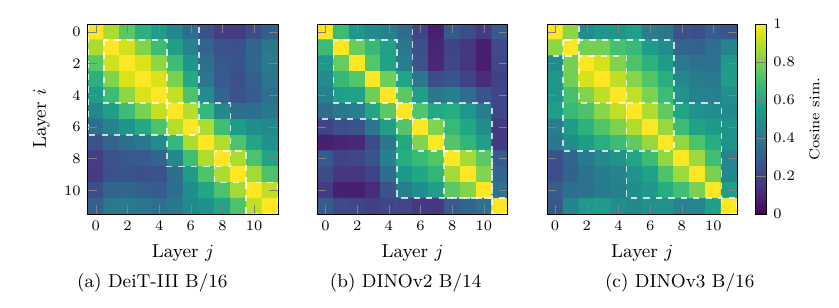}
    \vspace{-4pt}
    \caption{Pairwise cosine similarity between block outputs from ImageNet-1k reveals consistent block-diagonal structure across architectures and training paradigms, both for self-supervised (DINOv2, DINOv3) and supervised (DeiT-III) models. Contiguous groups of blocks produce redundant representations, which can be leveraged to produce more effective models. Possible blocks are shown in white.}
    \label{fig:modelsim}
\end{figure}

\paragraph{ViT operators.} At its core, a transformer $F_\Theta$ can be decomposed as a set of transformations
\begin{align}
  \label{eq:layerwise_vit}
  F_\Theta = f_{\theta_L} \circ \cdots \circ f_{\theta_1},
\end{align}
where $f_{\theta_\ell} = f(\plh; \theta_\ell)$ denotes a unique parametrisation for each layer $\ell$, and the full model parametrisation is given by $F_\Theta = F(\plh; \Theta)$ with $\Theta = (\theta_1, \dots, \theta_L)$.
Alternatively, \Cref{eq:layerwise_vit} can instead be formulated as contiguous block segments
\begin{align}
  F_\Theta = F_{\Theta_m} \circ \cdots \circ F_{\Theta_1},
\end{align}
where each subnetwork $F_{\Theta_j}$ for $1 \leq j \leq m$ contains layers with similar internal dynamics, \ie, some form of \textit{block redundancy}.
We begin by separating the phenomenon from the methods used to exploit it.

\paragraph{Block redundancy.}
Let $B_j = [s_j, e_j]$ be a contiguous block of length $k_j = e_j - s_j + 1$, where $1 \leq s_j < e_j \leq L$.
We say that block $B_j$ is $\varepsilon$-functionally-redundant if the layers within it implement highly similar transformations on the states visited by the model, such that
\begin{align}
  \max_{a, b \in B_j}
  \E_{x \sim \mathcal{D}} \Big[
      d\Big(
        f_{\theta_a}\big(h_{a-1}(x)\big),
        f_{\theta_b}\big(h_{b-1}(x)\big)
      \Big)
  \Big]
  \le \varepsilon,
\end{align}
where $h_{\ell}(x) = (f_{\theta_\ell} \circ \dots \circ f_{\theta_1})(x)$ denotes the intermediate hidden state after layer $\ell$, $d$ is a representation-space discrepancy measure, and $\mathcal{D}$ is a data distribution.
In simple terms, a phase is block-redundant when all layers behave similarly across depth.
Notably, \citet{jacobs2026block} provide an analysis under $d(x,y) = 1-\cos(x,y)$.
Empirical observations of the layer-to-layer distance similarity matrix
\begin{align}
  S_{a,b} = \E_{x \sim \mathcal{D}} \left[ d\Big(
        f_{\theta_a}\big(h_{a-1}(x)\big),
        f_{\theta_b}\big(h_{b-1}(x)\big)
      \Big) \right]
\end{align}
reveal distinct block-diagonal structures. 
These structures define \textit{blocks} $B_j$ which approximate the ideal \textit{phases} of computation, where the intra-block representational similarity remains exceptionally high, as shown in \Cref{fig:modelsim}.
This implies that $F_\Theta$ slowly integrates information over time to reach the final representations.

\paragraph{From redundancy to surrogate models.}
Once a block has been identified as functionally redundant, we can form a stronger and more constructive query.
Can an entire block be approximated by a simpler surrogate?
Let
\begin{align}
  F_{\Theta_j} = f_{\theta_{e_j}} \circ \cdots \circ f_{\theta_{s_j}}
\end{align}
denote the subnetwork implemented in block $B_j$ with parameters $\Theta_j = (\theta_{s_j}, \dots \theta_{e_j})$.
We then seek a \textit{surrogate mapping} $g_j$ such that
\begin{align}
  \E_{x \sim \mathcal{D}} \Big[
      d\Big(
        F_{\Theta_j}\big(h_{s_j-1}(x)\big),
        g_{j}\big(h_{s_j-1}(x)\big)
      \Big)
  \Big]
  \le \varepsilon.
\end{align}
The key distinction from previous approaches~\citep{jacobs2026block,cannistraci2026toast} and our proposed \shortmethod lies not in whether such a surrogate exists, but rather on the underlying assumptions motivating the choice of $g_j$.

\subsection{Recurrent and Non-Mixing Surrogates}

\paragraph{The Block Recurrent Hypothesis.}
\citet{jacobs2026block} operationalises block redundancy through a specific structural choice for $g_j$, and asks whether the surrogate can be chosen to be a repeated application of a single parameter-tied operator.
Under this view, one writes
\begin{align}
  g_j = f_{\theta_j^*}^{k_j} = \underbrace{f_{\theta_j^*} \circ \cdots \circ f_{\theta_j^*}}_{k_j \text{ times}} = f^{k_j}(\plh; \theta^*_j),
\end{align}
such that the same parametrisation $\theta_j^*$ is reused $k_j$ times inside block $j$.
Their proposed Raptor method optimises the parametrisation $\theta^*_j$ via
\begin{align}
  \theta^*_j = \argmin_\phi \, \E_{x \sim \mathcal D}\big\Vert F_{\Theta_j}\big(h_{s_j-1}(x)\big) - f^{k_j}_\phi\big(h_{s_j-1}(x)\big) \big\Vert_2^2,
\end{align}
which fine-tunes an existing model via a recurrent objective, noting that Raptor performs this optimisation in two separate training stages.

While Raptor demonstrates that contiguous phases can be approximated by shared-weight layers, we argue that recurrence is an overly specific interpretation of a broader geometric phenomenon.
In a residual architecture, small functional updates along a locally flat computational phase naturally induce block-diagonal similarity.
A sequence of independent layers traversing such a phase can therefore yield a similar signature as a recurrent operator, without implementing repeated dynamics or requiring identical weights.
From this perspective, recurrence is a useful parameter-sharing regulariser imposed on redundant phases, but not the fundamental mechanism generating them.
And because it still executes all $k_j$ transformations, computations at inference remain largely intact.

\paragraph{Non-mixing Surrogates.}
TOAST~\citep{cannistraci2026toast} takes the opposite approach to BRH\@. 
Rather than imposing recurrence, it bypasses redundant phases entirely by fitting a closed-form linear map between their endpoints. 
Similarly, NOSE~\citep{lin2024mlp} replaces specific ViT layers exclusively with feed-forward networks, adding more expressivity at the cost of more compute.
While both reduce parameters and raw compute without retraining, this strategy also limits the expressivity of the surrogate $g_j$.

A key property of attention operators is their capability for \textit{dynamic token-mixing}, which provides a high degree of expressivity~\citep{cordonnier2020Oconvatt,perez2021turingatt}.
A linear map can somewhat preserve proximity, but it cannot meaningfully reproduce token interactions between layers in identified blocks. 
While TOAST and NOSE demonstrate that entire phases can be fused using strictly less-expressive operators, they produce models that cannot retain the expressivity of their original components.

\subsection{\method}
\label{subsec:formalizing_collapse}

\shortmethod looks to resolve the tension between surrogate expressivity and inference efficiency.
We hypothesise that if a contiguous block $B_j = [s_j, e_j]$ approximates a single ideal computational phase, the intermediate representations, $h_{s_j}(x), \dots, h_{e_j-1}(x)$, are not strict necessities, but transitional micro-steps across a locally flat manifold.
Therefore, the surrogate $g_j$ can be chosen as a single, non-recurrent operator $f_{\phi_j}$ that computes the phase directly via
\begin{equation}
  \label{eq:twt_main}
  h_{e_j}(x) \approx f_{\phi_j}\big(h_{s_j-1}(x)\big).
\end{equation}
Crucially, to retain dynamic token-mixing, we restrict $f_{\phi_j}$ to the standard architecture of a ViT layer: a multi-head attention mechanism and a feed-forward network.
In replacing a redundant block with exactly one layer, \shortmethod fuses the computational phase into a single step, circumventing the iterative inference cost retained by BRH~\citep{jacobs2026block}.

This formalisation fundamentally alters the algorithmic interpretation of ViT depth.
It implies that deep networks do not strictly require $k_j$ distinct attention passes to incrementally route information within a phase.
Instead, a single, optimally parameterised attention and feed-forward pass possesses sufficient representational capacity to execute the entire spatial and channel-wise reorientation required for that ideal phase.

\section{\method Surrogate Discovery}
\label{sec:surrogate}
% \section{Phase-Collapse Distillation: The Proposed Method}
% \label{sec:method}

\begin{figure}[tb]
\centering
\includegraphics[width=\linewidth]{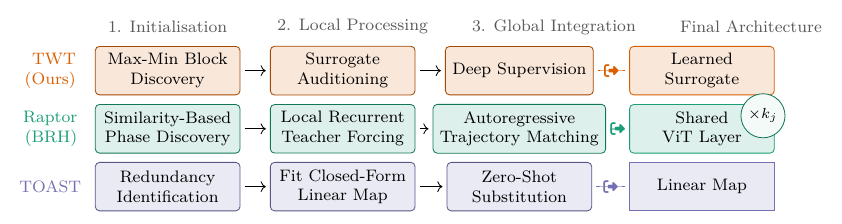}
% \vspace{5pt}
\caption{\textbf{Pipeline and architectural comparison of phase compression methods.}
% This hybrid diagram contrasts the optimisation procedures (left) against the resulting structural mechanisms (right) used to replace a redundant computational phase. (Top) TWT collapses the entire phase into a single, highly expressive surrogate layer via a unified, single-stage deep supervision pipeline targeting only block boundaries. (Middle) Raptor (BRH)~\citep{jacobs2026block} relies on a disjointed two-stage training process to stabilise recurrent gradients, yielding a parameter-tied layer that mimics intermediate steps and still must iterate at inference, incurring the same inference cost as the unpruned model. (Bottom) TOAST~\citep{cannistraci2026toast} avoids gradient-based training by fitting a closed-form linear map, but drops the standard transformer topology, completely sacrificing dynamic token-mixing expressivity.
TWT collapses each phase into one learned ViT surrogate using auditioning and single-stage deep supervision. Raptor (BRH)~\citep{jacobs2026block} recurrently reuses a shared ViT layer and therefore retains the original inference cost, whereas TOAST~\citep{cannistraci2026toast} fits a closed-form linear map without dynamic token mixing.
}
\label{fig:twt_process}
\end{figure}

To operationalise block redundancy, we propose a pruning and distillation framework that condenses contiguous blocks of learned operators into a single algorithmic step reflecting the ideal phase.
Unlike the BRH, which requires a surrogate operator $g_j = f_{\theta_j^*}^{k_j}$ to iteratively unroll $k_j$ times to mimic a target block $B_j$, our method approximates a block's computational phase with a single application of $f_{\phi_j}$ without recurrent iteration, as formalised in \Cref{eq:twt_main}.
We achieve this through a three-step process: (1)~dynamic block discovery, (2)~optimal candidate initialisation, and (3)~distillation via single-stage deep supervision.
We depict this process and contrast it against the literature in \Cref{fig:twt_process}.

\subsection{Block Discovery via Max-Min Dynamic Programming}
\label{subsec:phase_discovery}

Our first objective is to identify contiguous subsets of layers that form $\varepsilon$-functionally-redundant blocks.
Let $F_\Theta$ be a trained Vision Transformer with $L$ layers.
We compute the discrepancy matrix $S \in \mathbb{R}^{L \times L}$, where $S_{a,b}$ is the expected discrepancy $d$ of the token representations $h_a(x)$ and $h_b(x)$ over a representative calibration set $\mathcal{D}$, as introduced in \Cref{sec:pch}. 
In practice, we let $d$ be the cosine distance between intermediate activations.

We frame block discovery as a partitioning problem over depth.
We seek a set of contiguous blocks $\mathcal{P} = \{B_1, \dots, B_m\}$ that cover the entire network, where each $B_j = [s_j, e_j]$ and $s_{j+1} = e_j + 1$.
A valid partition must satisfy a maximum intra-block discrepancy threshold $\varepsilon$, ensuring that the boundary discrepancy satisfies $S_{s_j, e_j} \le \varepsilon$ for all $j$.

To prevent catastrophic degradation, we employ a min-max dynamic programming approach.
We first minimise the total number of blocks $m$, effectively maximising compression.
To break ties among equally minimal partitions, we select the partition that minimises the worst-case discrepancy among its blocks.
Formally, we optimise
\begin{align}
\min_{\mathcal{P}} \left( m, \max_{j \in \{1 \dots m\}} S_{s_j, e_j} \right) \quad \text{s.t.} \quad S_{s_j, e_j} \le \varepsilon, \forall j.
\end{align}
This yields a deterministic merge plan where each block $B_j$ corresponds to a single ideal phase to be captured by a surrogate operator.
We provide a sensitivity analysis in \appref{supp:sensitivity} showing how varying the threshold $\varepsilon$ controls the granularity of this partition.

\subsection{Macro-Step Initialisation via Auditioning}
\label{subsec:auditioning}

Instead of initialising a surrogate operator $f_{\phi_j}$ randomly, we exploit the parameters the teacher has already learned.
For a given block $B_j = [s_j, e_j]$, we generate a pool of candidate operators.
This pool, denoted as $\mathcal{C}_j = \{ f_{\theta_\ell}: \ell \in B_j \} \cup \{ \bar{f}_j \}$, includes each individual layer from the teacher's block, as well as an averaged operator $\bar{f}_j$ constructed by averaging the weight matrices of all layers in the span (while keeping normalisation parameters isolated).
We evaluate each candidate on a small calibration batch to minimise the local mapping error
\begin{align}
f^*_{\phi_j} = \argmin_{f \in \mathcal{C}_j} \E_{x \sim \mathcal{D}} \left[ \left\Vert f(h_{s_j-1}(x)) - h_{e_j}(x) \right\Vert_2^2 \right].
\end{align}
This \textit{auditioning} process identifies the operator best naturally positioned to execute the macro-step, significantly stabilising early training dynamics.

We provide an ablation study in \appref{supp:ablation} showing that the auditioned candidate generally outperforms the worst candidate, supporting the value of auditioning for downstream transfer.

\subsection{Distillation via Deep Supervision}
\label{subsec:distillation}

With the collapsed architecture initialised, we distil the teacher into the student without recurrent rollouts or linear surrogates.
Instead of the dual-stage training procedure used in Raptor~\citep{jacobs2026block}, we stitch the $K$ surrogate blocks together into a unified student model $F_{\Phi}$ and optimise it end-to-end.

Let $\mathcal{B}=\{(s_j,e_j)\}_{j=1}^{K}$ denote the merge plan, where student block $j$ replaces teacher blocks $s_j,\dots,e_j$.
For an input image $x$, let $h_j(x;\Phi)$ be the output of student block $j$, and let $h_{e_j}(x;\Theta)$ be the output of the last teacher block in the corresponding teacher segment.
Let $F_{\Phi}(x)$ and $F_{\Theta}(x)$ denote the final pre-classification backbone features of the student and teacher, respectively.
To gradually introduce deeper supervision terms, we use a staggered cosine schedule over normalised training time $\tau \in [0,1]$:
\begin{align}
\omega_j(\tau)=
\begin{cases}
\frac{1-\cos\!\left(\pi \tau / \alpha_j\right)}{2}, & 0 \le \tau < \alpha_j,\\[4pt]
1, & \alpha_j \le \tau,
\end{cases}
\end{align}
where $\alpha_j \in (0,1]$ is the activation time assigned to block $j$, with shallower blocks activated earlier and deeper blocks later.
The training objective is
\begin{align}
\mathcal{L}(x,\tau)
=
\sum_{j=1}^{K}
\omega_j(\tau)\,
\mathrm{MSE}\!\left(h_j(x;\Phi),\, h_{e_j}(x;\Theta)\right)
+
\omega_{K+1}(\tau)\,
\mathrm{MSE}\!\left(F_{\Phi}(x),\, F_{\Theta}(x)\right).
\end{align}
Thus, the student is supervised both at intermediate block outputs and at the final representation, with deeper losses introduced progressively over training.
All student parameters are optimised jointly using AdamW\@.

\begin{figure}[tb]
\centering

% \definecolor{sdinov2}{RGB}{76,114,176}
% \definecolor{sdinov3}{RGB}{221,132,82}
% \definecolor{sdeit3}{RGB}{85,168,104}
\colorlet{sdinov2}{Dark2-A}
\colorlet{sdinov3}{Dark2-B}
\colorlet{sdeit3}{Dark2-C}

\includegraphics[width=0.95\linewidth]{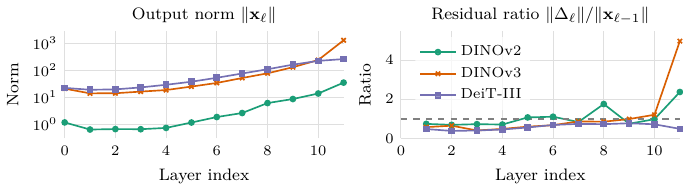}
\caption{\textbf{Norm growth across layers} (patch tokens, averaged over 50k ImageNet samples). \emph{Left:} output norms grow exponentially in later layers, with DINOv3 exhibiting a $63\times$ increase from the first to the last layer. \emph{Right:} the residual-to-output ratio exceeds $1$ in late layers for the self-supervised models, meaning the residual addition is larger than the accumulated representation. This scale imbalance motivates introducing learnable LayerScale parameters to stabilise convergence during distillation.}
\label{fig:normgrowth}
\end{figure}

\paragraph{Layer Scaling.} While most pre-trained models use LayerScale~\citep{touvron2021deepervit} during initial optimisation, these parameters are not always included in pre-trained checkpoints.
Our experiments indicate that several pre-trained models have a marked increase in norms for later layers, particularly in natural image models, as seen in \Cref{fig:normgrowth}.
To improve convergence, we explicitly introduce LayerScale, initialised as identity to allow the model to more easily adapt norms to dropped layers in block fusion. 
We find that a single scalar parameter typically suffices to improve convergence during fitting. 
These are fused with existing parameters in the final model weights.

\section{Geometric Evidence of Block Collapse}
\label{sec:analyses}

The formulation in \Cref{subsec:formalizing_collapse} posits that a single learned surrogate layer can approximate a contiguous redundant block. 
To test this hypothesis, we examine how collapse alters the representation space.
If intermediate layers within a block primarily refine a shared computational phase rather than introducing a new representational stage, the pruned student network should preserve the teacher's coarse trajectory at block boundaries. 
We quantify this geometric alignment by evaluating the intrinsic dimension (ID) of the token representations across depth using the Two-NN estimator~\cite{facco2017estimating}.

We compare an unpruned teacher model using the H0-mini~\cite{filiot2025distilling} backbone for histopathology images and DINOv2~\cite{oquab2024dinov2} for natural images against the corresponding pruned student networks. 
To capture the full scope of the representational flow, we track the ID of the global class token, the spatially averaged patch tokens, and the raw, unpooled flattened patch token sequence.

\begin{figure}[tb]
\centering

% Define Seaborn-like colours
% \definecolor{sblue}{RGB}{76,114,176}
% \definecolor{sorange}{RGB}{221,132,82}
\colorlet{sblue}{Dark2-A}
\colorlet{sorange}{Dark2-B}
\includegraphics[width=\linewidth]{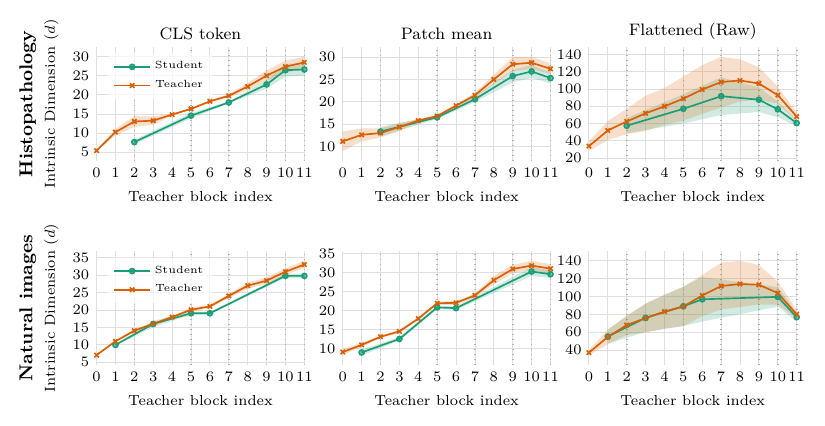}
\caption{\textbf{Intrinsic dimension (ID) trajectory comparison.} ID ($d$) estimated via the Two-NN estimator across H0-mini (top) and DINOv2 (bottom). We compare the IDs of: the global class (CLS) token (left), the spatially averaged patch tokens (centre), and the raw, unpooled sequence of all tokens flattened into a single vector (right). In the histopathology setting, the student closely tracks the teacher's manifold trajectory while exhibiting markedly lower variance. This regularisation effect is not present in natural images, where student and teacher variance remain comparable, suggesting that the variance reduction is specific to the domain-specialised distillation regime. Shaded bands denote $\pm 1$ standard deviation across random seeds and data augmentations.}
\label{fig:alignment}
\end{figure}

\subsection{Intrinsic Dimension and Smoothing}
\label{subsec:id_analysis}

The raw, flattened token representation reveals geometric changes that are less apparent after pooling.
Across block boundaries, the student follows the same broad ID trajectory as the unpruned teacher, indicating that the surrogate layers preserve the coarse representational progression of the original network. 
At the same time, the student consistently has a lower ID in the histopathology setting, particularly for the flattened patch-token representation in the rightmost panel of \Cref{fig:alignment}.
This pattern suggests that block collapse removes part of the high-dimensional variation introduced by the teacher's intermediate layers while retaining the endpoint geometry.

\subsection{Variance Reduction Under Stochastic Augmentation}
\label{subsec:augmentation_variance}

We next evaluate whether the lower ID observed in the student is accompanied by greater stability under input perturbations. For each input, we apply stochastic spatial and colour augmentations and measure the standard deviation of the resulting token representations at block boundaries. The shaded regions in \Cref{fig:alignment} represent the variance across these augmentation passes, showing larger variation for the unpruned teacher at several intermediate boundaries.
We quantify this instability in \Cref{tab:id_variance}.

\begin{table}[tb]
\centering
\footnotesize
\caption{Intrinsic Dimension (ID) and Standard Deviation (STD) of Flattened Tokens at Block Boundaries (Overall Combined: Clean + Augmented).
The unpruned teacher is H0-mini, while the block-collapsed student is the pruned H0-mini obtained by replacing the discovered blocks with 6 surrogate layers.
The student network shows consistently lower ID and reduced variance, indicating lower sensitivity to the sampled augmentations.}
\label{tab:id_variance}
\begin{tblr}{
  colspec = {
  c
  Q[c, si={table-format=3.1(2.1), uncertainty-mode=separate}]
  Q[c, si={table-format=2.1(3), uncertainty-mode=separate}]
  c
  },
  row{1,2} = {guard},
  cell{3-8}{4} = {font=\bfseries}
}
\toprule
Block Boundary & Unpruned Teacher & Block-Collapsed Student & STD Reduction \\
(Layer Index)  & ID ($\mu \pm \sigma$)    & ID ($\mu \pm \sigma$)          & (\%)      \\
\midrule
2  & 70.1(21.4)  & 58.3(11.2) & 47.6\% \\
5  & 89.4(26.1)  & 77.1(14.8) & 43.2\% \\
7  & 109.8(29.5) & 91.2(17.5) & 40.6\% \\
9  & 106.5(27.2) & 87.4(16.9) & 37.8\% \\
10 & 92.3(24.8)  & 76.5(14.1) & 43.1\% \\
11 & 68.4(12.5)  & 60.2(6.3)  & 49.6\% \\
\bottomrule
\end{tblr}
\end{table}

As shown in \Cref{tab:id_variance}, the block-collapsed student demonstrates a substantial reduction in representational standard deviation---ranging from 37.8\% to 49.6\% across all boundaries compared to the teacher network. 
This indicates that the collapsed student is less sensitive to augmentations at each measured boundary. 
Since the student still follows the teacher's coarse ID trajectory, the lower variance suggests that the surrogate layers preserve the phase boundaries while reducing augmentation-sensitive variation along the intermediate path.

This can be interpreted as evidence for the claim in \Cref{sec:surrogate}. 
Intermediate layers within a redundant block are not contributing distinct computational stages, but instead accumulate stochastic variation as they traverse a locally flat region of the representation manifold. 
By removing these transitional steps, \shortmethod can provide a more compact, stable representation with less sensitivity to perturbations by augmentation.
The downstream results in \Cref{tab:mil_results_h0_mini,tab:mil_results_hibou_b} confirm that this source of variation carries little-to-no task-relevant signal and is likely noise introduced by over-provisioned depth.

\section{Experiments}
\label{sec:experiments}
We evaluate \shortmethod across two distinct visual domains: natural images and histopathology Whole-Slide Image (WSI) classification. 
Our primary objective is to demonstrate that collapsing redundant blocks not only reduces parameter counts and computation but also does so with little to no performance loss.
We compare \shortmethod against Raptor~\citep{jacobs2026block} via BRH and two state-of-the-art depth pruning methods: WDPruning \cite{yu2022width} and NOSE \cite{lin2024mlp}, where each method is applied to a baseline model (reported as such in each result for reference).

\subsection{Histopathology Models}
\label{subsec:histopathology}

Histopathology models process gigapixel whole-slide images (WSIs), where inference cost scales with the number of retained tissue tiles per slide. We, therefore, evaluate whether \shortmethod preserves slide-level performance after reducing the depth of two domain-specific DINOv2-style backbones: H0-mini~\cite{filiot2025distilling} and Hibou-B~\cite{nechaev2024hibou}.

We use PANDA training tiles for histopathology phase discovery and backbone compression. We then freeze each pruned or unpruned backbone and precompute tile embeddings for all downstream cohorts. Each WSI forms a bag of foreground tile embeddings, and we train slide-level MIL heads on these frozen bags. We evaluate two MIL aggregators, ABMIL~\cite{ilse2018attention} and TransMIL~\cite{shao2021transmil}, to test whether the compressed features remain useful across different slide-level pooling mechanisms.

For prostate cancer grading, we train MIL heads on PANDA~\cite{bulten2022artificial} using the split from Song et al.~\cite{song2024morphological}. We select checkpoints by PANDA validation QWK and report QWK ($\singletimes100$) on PANDA test and TCGA-PRAD~\cite{zuley2016radiology}. For breast metastasis detection, we train MIL heads on CAMELYON17~\cite{litjens20181399} using corrected binary labels from Ling et al.~\cite{ling2025comprehensive}. We select checkpoints by the 5-epoch moving average of CAMELYON17 validation accuracy and report slide-level accuracy on CAMELYON17 test and CAMELYON16~\cite{bejnordi2017diagnostic}. We provide tiling, tissue filtering, label processing, and optimisation details in \appref{supp:histopathology}.

\begin{table*}[tb]
    \scriptsize
    \caption{Histopathology MIL results for H0-mini. We report QWK ($\singletimes100$) for prostate grading and accuracy for breast metastasis detection. Prostate MIL heads train on PANDA and test on PANDA/TCGA-PRAD; metastasis MIL heads train on CAMELYON17 and test on CAMELYON17/CAMELYON16. Deltas compare against the uncompressed backbone with the same MIL head.}
    \label{tab:mil_results_h0_mini}
    \centering
    \adjustbox{max width=\textwidth, keepaspectratio}{%
    \begin{tblr}{
      % Added the 'c' for FLOPs. Total columns is now 13.
      colspec = {c l c c c *{4}{lr}}, 
      % Shifted from 5-12 to 6-13 to encompass all metric data and deltas
      column{6-13} = {mode=math},
      % Shifted delta styling columns from {6,8,10,12} to {7,9,11,13}
      column{7,9,11,13} = {font=\tiny, leftsep=-3pt},
      row{1,2} = {mode=text},
      row{9,10,17,18} = {bg=highlight},
      % Updated cell references to target the correct color columns (+1 to column indices)
      cell{4-6,10,12,17,18}{7} = {fg=red},
      cell{7-9,13-16}{7} = {fg=teal},
      cell{4-6,12-14}{9} = {fg=red},
      cell{7-10,15-18}{9} = {fg=teal},
      cell{12}{11} = {fg=red},
      cell{4-10,13-18}{11} = {fg=teal},
      cell{4-10,12-14,16-18}{13} = {fg=red},
      cell{15}{13} = {fg=teal},
    }
    \toprule
    \SetCell[c=5]{c} & & & & & \SetCell[c=4]{c} Prostate (QWK) & & & & \SetCell[c=4]{c} Breast/Lymph (Acc) & & & \\
    \cmidrule[r]{6-9} 
    \cmidrule[r]{10-13}
    Agg. & Method & Layers & Params (M) & FLOPs (G) & 
     \SetCell[c=2]{l}{PANDA Test} & & \SetCell[c=2]{l}{TCGA-PRAD} & & \SetCell[c=2]{l}{CAM17 Test} & & \SetCell[c=2]{l}{CAM16} & \\
    \midrule
    \SetCell[r=8]{c} \rotatebox{90}{ABMIL} & Baseline & 12 & 85.74 & 23.56 & \SetCell[c=2]{l} 93.77 \pm 0.16 & & \SetCell[c=2]{l} 69.14 \pm 1.22 & & \SetCell[c=2]{l} 88.35 \pm 0.72 & & \SetCell[c=2]{l} 97.31 \pm 0.23 & \\
    & NOSE & 12 & 57.36 & 14.91 & 85.40 \pm 0.26 & -8.38 & 42.56 \pm 2.99 & -26.58 & 89.24 \pm 0.98 & +0.89 & 81.30 \pm 1.60 & -16.01 \\
    & WDPruning & 5 & 42.40 & \phantom{0}9.37 & 93.44 \pm 0.26 & -0.33 & 64.65 \pm 1.59 & -4.49 & 91.08 \pm 1.30 & +2.73 & 91.09 \pm 2.73 & -6.22 \\
    &  & 4 & 35.31 & \phantom{0}7.52 & 93.65 \pm 0.21 & -0.13 & 55.52 \pm 0.28 & -13.62 & 91.59 \pm 0.48 & +3.24 & 90.26 \pm 0.75 & -7.05 \\
    & Raptor & 5 & 36.30 & 23.57 & 94.23 \pm 0.22 & +0.45 & 69.30 \pm 0.59 & +0.16 & 89.28 \pm 0.92 & +0.93 & 96.94 \pm 0.22 & -0.36 \\
    &  & 4 & 29.17 & 23.57 & 93.97 \pm 0.13 & +0.20 & 69.38 \pm 0.37 & +0.25 & 89.32 \pm 0.68 & +0.97 & 96.94 \pm 0.34 & -0.36 \\
    & \shortmethod & 5 & 36.11 & \phantom{0}9.36 & 93.79 \pm 0.25 & +0.02 & 70.08 \pm 1.08 & +0.94 & 90.46 \pm 0.43 & +2.12 & 96.73 \pm 0.39 & -0.57 \\
    &  & 4 & 29.02 & \phantom{0}7.51 & 93.71 \pm 0.28 & -0.06 & 69.69 \pm 0.55 & +0.55 & 89.70 \pm 1.75 & +1.36 & 95.49 \pm 0.70 & -1.81 \\
    \midrule
    \SetCell[r=8]{c} \rotatebox{90}{TransMIL} & Baseline & 12 & 85.74 & 23.56 & \SetCell[c=2]{l} 94.57 \pm 0.41 & & \SetCell[c=2]{l} 56.02 \pm 6.21 & & \SetCell[c=2]{l} 89.11 \pm 0.71 & & \SetCell[c=2]{l} 96.27 \pm 0.98 & \\
    & NOSE & 12 & 57.36 & 14.91 & 89.93 \pm 0.57 & -4.64 & 49.15 \pm 2.22 & -6.87 & 88.65 \pm 1.43 & -0.46 & 81.23 \pm 0.93 & -15.04 \\
    & WDPruning & 5 & 42.40 & \phantom{0}9.37 & 94.62 \pm 0.30 & +0.05 & 51.29 \pm 7.10 & -4.73 & 90.61 \pm 1.31 & +1.50 & 90.65 \pm 1.27 & -5.62 \\
    &  & 4 & 35.31 & \phantom{0}7.52 & 94.66 \pm 0.50 & +0.09 & 38.19 \pm 3.98 & -17.83 & 89.97 \pm 1.32 & +0.86 & 88.20 \pm 1.93 & -8.07 \\
    & Raptor & 5 & 36.30 & 23.57 & 94.62 \pm 0.59 & +0.04 & 57.22 \pm 5.63 & +1.20 & 90.84 \pm 0.64 & +1.73 & 96.84 \pm 0.50 & +0.57 \\
    &  & 4 & 29.17 & 23.57 & 94.64 \pm 0.39 & +0.07 & 57.73 \pm 8.90 & +1.71 & 90.04 \pm 0.50 & +0.93 & 96.01 \pm 1.41 & -0.26 \\
    & \shortmethod & 5 & 36.11 & \phantom{0}9.36 & 94.49 \pm 0.67 & -0.08 & 61.89 \pm 3.56 & +5.87 & 90.13 \pm 0.71 & +1.02 & 95.29 \pm 0.67 & -0.98 \\
    &  & 4 & 29.02 & \phantom{0}7.51 & 94.29 \pm 0.59 & -0.28 & 59.11 \pm 4.13 & +3.09 & 90.93 \pm 0.86 & +1.82 & 95.23 \pm 0.59 & -1.04 \\
    \bottomrule
    \end{tblr}
    }%
    % \vspace{-10pt}
\end{table*}

\begin{table*}[tb]
    \scriptsize
    \caption{Histopathology MIL results for Hibou-B. We report QWK ($\singletimes100$) for prostate grading and accuracy for breast metastasis detection. Prostate MIL heads train on PANDA and test on PANDA/TCGA-PRAD; metastasis MIL heads train on CAMELYON17 and test on CAMELYON17/CAMELYON16. Deltas compare against the uncompressed backbone with the same MIL head.}
    \label{tab:mil_results_hibou_b}
    \centering
    \adjustbox{max width=\textwidth, keepaspectratio}{%
    \begin{tblr}{
      % Added the 'c' for FLOPs. Total columns is now 13.
      colspec = {c l c c c *{4}{lr}}, 
      % Shifted from 5-12 to 6-13 to encompass all metric data and deltas
      column{6-13} = {mode=math},
      % Shifted delta styling columns from {6,8,10,12} to {7,9,11,13}
      column{7,9,11,13} = {font=\tiny, leftsep=-3pt},
      row{1,2} = {mode=text},
      row{9,10,17,18} = {bg=highlight},
      % Updated cell references to target the correct color columns (+1 to column indices)
      cell{4-6,8-10,12,14,18}{7} = {fg=red},
      cell{7,13,15-17}{7} = {fg=teal},
      cell{4,6,10,12-14,17,18}{9} = {fg=red},
      cell{5,7-9,15,16}{9} = {fg=teal},
      cell{5,6,8,10,12,14-16}{11} = {fg=red},
      cell{4,7,9,13,17,18}{11} = {fg=teal},
      cell{4-6,12-16,18}{13} = {fg=red},
      cell{7-10,17}{13} = {fg=teal},
    }
    \toprule
    \SetCell[c=4]{c} & & & & & \SetCell[c=4]{c} Prostate (QWK) & & & & \SetCell[c=4]{c} Breast/Lymph (Acc) & & & \\
    \cmidrule[r]{6-9} 
    \cmidrule[r]{10-13} 
    Agg. & Method & Layers & Params (M) & FLOPs (G) & 
    \SetCell[c=2]{l}{PANDA Test} & & \SetCell[c=2]{l}{TCGA-PRAD} & & \SetCell[c=2]{l}{CAM17 Test} & & \SetCell[c=2]{l}{CAM16} & \\
    \midrule
    \SetCell[r=8]{c} \rotatebox{90}{ABMIL} & Baseline & 12 & 85.74 & 23.56 & \SetCell[c=2]{l} 93.70 \pm 0.24 & & \SetCell[c=2]{l} 66.27 \pm 1.11 & & \SetCell[c=2]{l} 90.25 \pm 0.87 & & \SetCell[c=2]{l} 90.52 \pm 2.48 & \\
    & NOSE & 12 & 57.36 & 14.91 & 86.49 \pm 0.21 & -7.21 & 51.02 \pm 2.15 & -15.26 & 90.47 \pm 0.84 & +0.21 & 83.94 \pm 0.99 & -6.58 \\
    & WDPruning & 4 & 35.31 & \phantom{0}7.94 & 93.64 \pm 0.15 & -0.07 & 66.55 \pm 0.89 & +0.28 & 89.66 \pm 1.58 & -0.59 & 73.99 \pm 6.77 & -16.53 \\
    &  & 3 & 28.22 & \phantom{0}5.99 & 92.96 \pm 0.46 & -0.75 & 64.66 \pm 0.98 & -1.61 & 88.14 \pm 0.00 & -2.11 & 65.54 \pm 0.00 & -24.98 \\
    & Raptor & 4 & 29.17 & 23.57 & 93.79 \pm 0.33 & +0.09 & 67.13 \pm 0.65 & +0.86 & 90.97 \pm 1.06 & +0.72 & 94.82 \pm 1.56 & +4.30 \\
    &  & 3 & 22.04 & 23.57 & 93.57 \pm 0.29 & -0.14 & 66.47 \pm 0.42 & +0.19 & 88.69 \pm 1.59 & -1.56 & 95.60 \pm 0.55 & +5.08 \\
    & \shortmethod & 4 & 29.02 & \phantom{0}7.93 & 93.27 \pm 0.32 & -0.44 & 67.77 \pm 0.67 & +1.50 & 90.25 \pm 1.03 & +0.00 & 96.16 \pm 0.28 & +5.64 \\
    &  & 3 & 21.93 & \phantom{0}5.98 & 93.30 \pm 0.21 & -0.41 & 65.49 \pm 0.60 & -0.79 & 89.66 \pm 0.92 & -0.59 & 93.63 \pm 1.43 & +3.11 \\
    \midrule
    \SetCell[r=8]{c} \rotatebox{90}{TransMIL} & Baseline & 12 & 85.74 & 23.56 & \SetCell[c=2]{l} 94.13 \pm 0.44 & & \SetCell[c=2]{l} 57.32 \pm 7.11 & & \SetCell[c=2]{l} 89.32 \pm 0.65 & & \SetCell[c=2]{l} 92.80 \pm 2.99 & \\
    & NOSE & 12 & 57.36 & 14.91 & 88.82 \pm 0.36 & -5.31 & 53.88 \pm 2.05 & -3.44 & 89.19 \pm 0.75 & -0.13 & 83.99 \pm 1.44 & -8.81 \\
    & WDPruning & 4 & 35.31 & \phantom{0}7.94 & 94.36 \pm 0.35 & +0.23 & 45.27 \pm 5.86 & -12.04 & 90.04 \pm 0.81 & +0.72 & 79.48 \pm 3.69 & -13.32 \\
    &  & 3 & 28.22 & \phantom{0}5.99 & 93.38 \pm 0.51 & -0.75 & 51.66 \pm 3.77 & -5.66 & 87.97 \pm 2.27 & -1.35 & 67.31 \pm 9.44 & -25.49 \\
    & Raptor & 4 & 29.17 & 23.57 & 94.15 \pm 0.51 & +0.02 & 59.49 \pm 6.96 & +2.18 & 88.64 \pm 1.39 & -0.68 & 74.30 \pm 9.00 & -18.50 \\
    &  & 3 & 22.04 & 23.57 & 94.67 \pm 0.18 & +0.54 & 58.59 \pm 8.49 & +1.27 & 89.02 \pm 0.66 & -0.30 & 91.55 \pm 5.29 & -1.25 \\
    & \shortmethod & 4 & 29.02 & \phantom{0}7.93 & 94.65 \pm 0.29 & +0.52 & 55.47 \pm 2.70 & -1.85 & 91.02 \pm 1.12 & +1.70 & 93.26 \pm 2.36 & +0.46 \\
    &  & 3 & 21.93 & \phantom{0}5.98 & 94.02 \pm 0.07 & -0.11 & 54.58 \pm 3.12 & -2.74 & 90.42 \pm 0.98 & +1.10 & 91.09 \pm 2.84 & -1.71 \\
    \bottomrule
    % \SetCell[c=12]{l} \tdgg Hibou-B / Raptor underperforms, but is included for completeness. See the \hyperlink{paragraph:hibou_results}{Hibou-B results paragraph} for details. 
    \end{tblr}
    }%
    % \vspace{-5pt}
\end{table*}

\paragraph{Pruning Implementation Details.} We train pruned backbones using AdamW~\cite{loshchilov2019decoupled}, a cosine learning-rate decay~\cite{loshchilov2016sgdr}, and gradient clipping at norm 1.0. We use a peak learning rate of $3\times10^{-4}$. We employ colour jitter and Gaussian blur data augmentations during distillation. Notably, we omit both the auxiliary LayerScale parameters and the staggered cosine loss schedule for these histopathology models. Unlike natural image backbones, they do not exhibit late-layer norm explosion and optimise stably without these additions. We scale training budgets by student size: for H0-mini, we train NOSE and WDPruning for 2 epochs, Raptor for 3 epochs total (1 in Stage-1, 2 in Stage-2), and \shortmethod for 2 epochs (depth 4) or 1.5 epochs (depth 5). For Hibou-B, we use the same hyperparameters as for H0-mini across all methods, with 2 epochs for \shortmethod at both depths. We also experimented with extended training budgets on Hibou-B (up to 5 epochs for NOSE and WDPruning, and 3 epochs for Raptor); however, these extended runs yielded similar downstream outcomes, confirming that the performance bottlenecks are not due to insufficient training time.

\paragraph{H0-mini results.}
\Cref{tab:mil_results_h0_mini} shows that \shortmethod preserves prostate grading performance while reducing the active backbone size by more than half. The five-layer \shortmethod model uses $36.11$M parameters instead of $85.74$M and matches the uncompressed model on PANDA for both ABMIL and TransMIL. It also improves TCGA-PRAD QWK for both MIL heads, with gains of $+0.94$ and $+5.87$, respectively. The four-layer model further reduces the active parameter count to $29.02$M and remains close to the baseline on PANDA while still improving TCGA-PRAD. On CAMELYON17, \shortmethod improves accuracy across both MIL heads and both depths. CAMELYON16 shows a small drop relative to the baseline, but the pruned models remain within roughly two percentage points while using about one third of the active backbone parameters.

\paragraph{Hibou-B results.}\hypertarget{paragraph:hibou_results}{}
\label{paragraph:hibou_results}
\Cref{tab:mil_results_hibou_b} evaluates whether the same compression behaviour transfers to a second histopathology foundation model. 
\shortmethod again provides a strong parameter--performance trade-off: the four-layer model uses $29.02$M active parameters compared with $85.74$M for the full backbone, while remaining close to the baseline on PANDA and improving ABMIL performance on TCGA-PRAD. 
On CAMELYON, \shortmethod is particularly strong for the external CAMELYON16 evaluation, indicating that the collapsed backbone can retain transferable features across tissue type and dataset shift. 
Some competing pruned baselines, such as WDPruning and NOSE, are less stable in this setting, especially at aggressive compression levels, but our \shortmethod achieves similar parameter-performance trade-offs while fundamentally reducing inference compute.

\subsection{Natural Image Models}
\label{subsec:natimage}

Our experiments with natural images closely follow the pruning methodology from \Cref{subsec:histopathology}, with few exceptions. 
For natural images, we use a total of 8 epochs across all methods, and while training stabilises quite early, we see a more dramatic effect from using a staggered cosine schedule for the loss function than in histopathology.
Our experiments focus on ImageNet-1k~\citep{deng2009imagenet} for classification and ADE20k~\citep{zhou2019semantic} for segmentation.
Evaluation protocols follow the respective baselines~\citep{touvron2022deit,jose2024dinov2,simeoni2025dinov3}.

\begin{table*}[tb]
    \footnotesize
    \caption{
      Natural image classification results for base-capacity backbones. 
      We report top-1 accuracy at $224 \times 224$. 
      Deltas compare against the uncompressed backbone of the same model.
    }
    \label{tab:natimgcls}
    \centering
    \begin{tblr}{
      colspec = {c lcc *{4}{lr}},
      column{5-12} = {mode=math},
      column{8,10,12} = {font=\tiny, leftsep=-3pt, fg=red},
      column{6} = {font=\tiny, leftsep=-3pt, fg=teal},
      row{1,2} = {mode=text},
      row{9,10,17,18,25,26} = {bg=highlight},
      cell{7,8,15,16,23,24}{6} = {fg=red},
    }
    \toprule
    \SetCell[c=6]{c} & & & & & &
    \SetCell[c=6]{c} ImageNet-1k & & & & & \\
    \cmidrule[r]{7-12}
    Model & Method & Layers & Params (M) &
    \SetCell[c=2]{c} FLOPs (G) & &
    Val & & ReaL & & v2 & \\
    \midrule
    \SetCell[r=8]{c} \rotatebox{90}{DEiT-III \, B/16} & Baseline & 12 & 86.6 & 17.81 & &
    \SetCell[c=2]{l} \NatVal{deitv3_base}{in_val} & &
    \SetCell[c=2]{l} \NatVal{deitv3_base}{in_real} & &
    \SetCell[c=2]{l} \NatVal{deitv3_base}{in_v2} & \\
    & NOSE & 12 & 70.1 & 14.06 & -3.75 &
    \NatVal{deitv3_nose}{in_val} & \NatDelta{deitv3_nose}{in_val} & 
    \NatVal{deitv3_nose}{in_real} & \NatDelta{deitv3_nose}{in_real} & 
    \NatVal{deitv3_nose}{in_v2} & \NatDelta{deitv3_nose}{in_v2} \\
    & WDPruning & 6 & 45.0 & \phantom{0}8.92 & -8.89 & 
    \NatVal{deitv3_wd_6}{in_val} & \NatDelta{deitv3_wd_6}{in_val} & 
    \NatVal{deitv3_wd_6}{in_real} & \NatDelta{deitv3_wd_6}{in_real} & 
    \NatVal{deitv3_wd_6}{in_v2} & \NatDelta{deitv3_wd_6}{in_v2} \\
    &  & 5 & 38.2 & \phantom{0}7.43 & -10.38 &
    \NatVal{deitv3_wd_5}{in_val} & \NatDelta{deitv3_wd_5}{in_val} & 
    \NatVal{deitv3_wd_5}{in_real} & \NatDelta{deitv3_wd_5}{in_real} & 
    \NatVal{deitv3_wd_5}{in_v2} & \NatDelta{deitv3_wd_5}{in_v2} \\
    & Raptor & 6 & 39.4 & 17.81 & +0.00 &
    \NatVal{deitv3_raptor_6}{in_val} & \NatDelta{deitv3_raptor_6}{in_val} & 
    \NatVal{deitv3_raptor_6}{in_real} & \NatDelta{deitv3_raptor_6}{in_real} & 
    \NatVal{deitv3_raptor_6}{in_v2} & \NatDelta{deitv3_raptor_6}{in_v2} \\
    &  & 5 & 32.5 & 17.81 & +0.00 &
    \NatVal{deitv3_raptor_5}{in_val} & \NatDelta{deitv3_raptor_5}{in_val} & 
    \NatVal{deitv3_raptor_5}{in_real} & \NatDelta{deitv3_raptor_5}{in_real} & 
    \NatVal{deitv3_raptor_5}{in_v2} & \NatDelta{deitv3_raptor_5}{in_v2} \\
    & \shortmethod & 6 & 39.0 & \phantom{0}8.91 & -8.90 &
    \NatVal{deitv3_twt_6}{in_val} & \NatDelta{deitv3_twt_6}{in_val} & 
    \NatVal{deitv3_twt_6}{in_real} & \NatDelta{deitv3_twt_6}{in_real} & 
    \NatVal{deitv3_twt_6}{in_v2} & \NatDelta{deitv3_twt_6}{in_v2} \\
    &  & 5 & 32.1 & \phantom{0}7.42 & -10.39 &
    \NatVal{deitv3_twt_5}{in_val} & \NatDelta{deitv3_twt_5}{in_val} & 
    \NatVal{deitv3_twt_5}{in_real} & \NatDelta{deitv3_twt_5}{in_real} & 
    \NatVal{deitv3_twt_5}{in_v2} & \NatDelta{deitv3_twt_5}{in_v2} \\
    \midrule
    \SetCell[r=8]{c} \rotatebox{90}{DINOv2 \, B/14} & Baseline & 12 & 86.6 & 23.42 & &
    \SetCell[c=2]{l} \NatVal{dinov2_base}{in_val} & &
    \SetCell[c=2]{l} \NatVal{dinov2_base}{in_real} & &
    \SetCell[c=2]{l} \NatVal{dinov2_base}{in_v2} & \\
    & NOSE & 12 & 70.1 & 18.38 & -5.04 &
    \NatVal{dinov2_nose}{in_val} & \NatDelta{dinov2_nose}{in_val} & 
    \NatVal{dinov2_nose}{in_real} & \NatDelta{dinov2_nose}{in_real} & 
    \NatVal{dinov2_nose}{in_v2} & \NatDelta{dinov2_nose}{in_v2} \\
    & WDPruning & 6 & 45.0 & 11.72 & -11.70 &
    \NatVal{dinov2_wd_6}{in_val} & \NatDelta{dinov2_wd_6}{in_val} & 
    \NatVal{dinov2_wd_6}{in_real} & \NatDelta{dinov2_wd_6}{in_real} & 
    \NatVal{dinov2_wd_6}{in_v2} & \NatDelta{dinov2_wd_6}{in_v2} \\
    &  & 5 & 38.2 & \phantom{0}9.77 & -13.65 &
    \NatVal{dinov2_wd_5}{in_val} & \NatDelta{dinov2_wd_5}{in_val} & 
    \NatVal{dinov2_wd_5}{in_real} & \NatDelta{dinov2_wd_5}{in_real} & 
    \NatVal{dinov2_wd_5}{in_v2} & \NatDelta{dinov2_wd_5}{in_v2} \\
    & Raptor & 6 & 39.4 & 23.42 & +0.00 &
    \NatVal{dinov2_raptor_6}{in_val} & \NatDelta{dinov2_raptor_6}{in_val} & 
    \NatVal{dinov2_raptor_6}{in_real} & \NatDelta{dinov2_raptor_6}{in_real} & 
    \NatVal{dinov2_raptor_6}{in_v2} & \NatDelta{dinov2_raptor_6}{in_v2} \\
    &  & 5 & 32.5 & 23.42 & +0.00 &
    \NatVal{dinov2_raptor_5}{in_val} & \NatDelta{dinov2_raptor_5}{in_val} & 
    \NatVal{dinov2_raptor_5}{in_real} & \NatDelta{dinov2_raptor_5}{in_real} & 
    \NatVal{dinov2_raptor_5}{in_v2} & \NatDelta{dinov2_raptor_5}{in_v2} \\
    & \shortmethod & 6 & 39.0 & 11.71 & -11.71 &
    \NatVal{dinov2_twt_6}{in_val} & \NatDelta{dinov2_twt_6}{in_val} & 
    \NatVal{dinov2_twt_6}{in_real} & \NatDelta{dinov2_twt_6}{in_real} & 
    \NatVal{dinov2_twt_6}{in_v2} & \NatDelta{dinov2_twt_6}{in_v2} \\
    &  & 5 & 32.1 & \phantom{0}9.76 & -13.66 &
    \NatVal{dinov2_twt_5}{in_val} & \NatDelta{dinov2_twt_5}{in_val} & 
    \NatVal{dinov2_twt_5}{in_real} & \NatDelta{dinov2_twt_5}{in_real} & 
    \NatVal{dinov2_twt_5}{in_v2} & \NatDelta{dinov2_twt_5}{in_v2} \\
    \midrule
    \SetCell[r=8]{c} \rotatebox{90}{DINOv3 \, B/16} & Baseline & 12 & 86.6 & 17.82 & &
    \SetCell[c=2]{l} \NatVal{dinov3_base}{in_val} & &
    \SetCell[c=2]{l} \NatVal{dinov3_base}{in_real} & &
    \SetCell[c=2]{l} \NatVal{dinov3_base}{in_v2} & \\
    & NOSE & 12 & 70.1 & 14.07 & -3.75 &
    \NatVal{dinov3_nose}{in_val} & \NatDelta{dinov3_nose}{in_val} & 
    \NatVal{dinov3_nose}{in_real} & \NatDelta{dinov3_nose}{in_real} & 
    \NatVal{dinov3_nose}{in_v2} & \NatDelta{dinov3_nose}{in_v2} \\
    & WDPruning & 6 & 45.0 & \phantom{0}8.93 & -8.89 &
    \NatVal{dinov3_wd_6}{in_val} & \NatDelta{dinov3_wd_6}{in_val} & 
    \NatVal{dinov3_wd_6}{in_real} & \NatDelta{dinov3_wd_6}{in_real} & 
    \NatVal{dinov3_wd_6}{in_v2} & \NatDelta{dinov3_wd_6}{in_v2} \\
    &  & 5 & 38.2 & \phantom{0}7.44 & -10.38 &
    \NatVal{dinov3_wd_5}{in_val} & \NatDelta{dinov3_wd_5}{in_val} & 
    \NatVal{dinov3_wd_5}{in_real} & \NatDelta{dinov3_wd_5}{in_real} & 
    \NatVal{dinov3_wd_5}{in_v2} & \NatDelta{dinov3_wd_5}{in_v2} \\
    & Raptor & 6 & 39.4 & 17.82 & +0.00 &
    \NatVal{dinov3_raptor_6}{in_val} & \NatDelta{dinov3_raptor_6}{in_val} & 
    \NatVal{dinov3_raptor_6}{in_real} & \NatDelta{dinov3_raptor_6}{in_real} & 
    \NatVal{dinov3_raptor_6}{in_v2} & \NatDelta{dinov3_raptor_6}{in_v2} \\
    &  & 5 & 32.5 & 17.82 & +0.00 &
    \NatVal{dinov3_raptor_5}{in_val} & \NatDelta{dinov3_raptor_5}{in_val} & 
    \NatVal{dinov3_raptor_5}{in_real} & \NatDelta{dinov3_raptor_5}{in_real} & 
    \NatVal{dinov3_raptor_5}{in_v2} & \NatDelta{dinov3_raptor_5}{in_v2} \\
    & \shortmethod & 6 & 39.0 & \phantom{0}8.92 & -8.90 &
    \NatVal{dinov3_twt_6}{in_val} & \NatDelta{dinov3_twt_6}{in_val} & 
    \NatVal{dinov3_twt_6}{in_real} & \NatDelta{dinov3_twt_6}{in_real} & 
    \NatVal{dinov3_twt_6}{in_v2} & \NatDelta{dinov3_twt_6}{in_v2} \\
    &  & 5 & 32.1 & \phantom{0}7.43 & -10.39 &
    \NatVal{dinov3_twt_5}{in_val} & \NatDelta{dinov3_twt_5}{in_val} & 
    \NatVal{dinov3_twt_5}{in_real} & \NatDelta{dinov3_twt_5}{in_real} & 
    \NatVal{dinov3_twt_5}{in_v2} & \NatDelta{dinov3_twt_5}{in_v2} \\
    \bottomrule
    \end{tblr}
    % \vspace{-5pt}
\end{table*}

\paragraph{Classification.} \Cref{tab:natimgcls} shows that \shortmethod achieves competitive accuracy at a fraction of the inference cost.
At six layers, \shortmethod matches or closely tracks Raptor across all three backbones while operating at roughly half the FLOPs; on DINOv2 B/14, the six-layer \shortmethod model outperforms Raptor on all folds despite requiring only 11.71\, GFLOPs versus 23.42\, GFLOPs.
Interestingly, Raptor~\citep{jacobs2026block} provides marginal accuracy improvements on B/16 models, notably with no reduction in compute.
With fewer tokens, the attention matrix is smaller ($196\times 196$ vs.\ $256\times 256$), so each recurrent application achieves more complete global mixing relative to the total information content. 
A recurrent layer could have a smoother optimisation target simply because the token-mixing space has a lower dimensionality.

Compared to WDPruning, which operates at a near-identical compute budget, \shortmethod consistently matches or improves accuracy, suggesting that the phase-aware surrogate initialisation and deep supervision yield a more effective model than uniform depth pruning.
At five layers, all methods degrade more noticeably, reflecting compression that likely goes beyond the redundant phases into functionally distinct computation.
Nevertheless, \shortmethod remains within approximately one point of the baseline across all three tested models.

In contrast to the histopathology results (\Cref{tab:mil_results_h0_mini,tab:mil_results_hibou_b}), no method improves over the uncompressed baseline on natural images.
This is consistent with our observations on smoothing from \Cref{subsec:id_analysis}; ImageNet backbones were trained on this distribution, so the redundant phases are already well-tuned and collapsing them can at best preserve performance.
In histopathology, the backbone operates out-of-distribution relative to its pre-training data, and the intermediate iterative steps accumulate domain-irrelevant noise that block collapse actively removes.

\paragraph{Dense Tasks.}
In addition to instance-level classification, we evaluate semantic segmentation on ADE20k, comparing against baseline DINO-family models and Raptor~\citep{jacobs2026block}.
\Cref{tab:natimgseg} shows that dense prediction is sensitive to block approximation even when the full recurrent execution cost is retained.
Raptor preserves the original network's iterative computation but still incurs a modest drop in mIoU.
In comparison, \shortmethod reduces inference compute by approximately $50\%$, at the cost of an additional 1.3 and 1.7 mIoU on DINOv2 and DINOv3, respectively.

This result is consistent with the distinction made by \citet{jacobs2026block} between patch-token and instance-level dynamics.
Dense prediction depends directly on the spatial token field, whereas classification can remain robust when the instance representation is preserved.
A recurrent surrogate may therefore retain local patch-token evolution more faithfully because it maintains iterative token mixing.
The additional mIoU loss therefore reflects a trade-off between preserving patch-level trajectory structure and eliminating recurrent inference cost, suggesting that dense prediction may require less aggressive block collapse.
We discuss this trade-off further in \Cref{subsec:limitations}.

\begin{table*}[t]
    \footnotesize
    \caption{
        Semantic segmentation results on ADE20k.
        Deltas compare against the uncompressed baseline.
    }
    \label{tab:natimgseg}
    \centering
    \begin{tblr}{
        colspec = {c l c c r c c r},
        column{5} = {font=\tiny, leftsep=-3pt, fg=teal},
        column{8} = {font=\tiny, leftsep=-3pt, fg=red},
        column{4,5,7,8} = {mode=math},
        row{1} = {guard, mode=text},
        row{4,7} = {bg=highlight},
        cell{3,6}{5} = {fg=red},
    }
        \toprule
        Model & Method & Layers & \SetCell[c=2]{c} FLOPs (G) & & Size &
        \SetCell[c=2]{c} mIoU & \\
        \midrule
        \SetCell[r=3]{c} \rotatebox{90}{\scriptsize DINOv2}
        & Baseline     & 12 & \SetCell[c=2]{l} 147.4 & & 518 $\times$ 518 &
        \SetCell[c=2]{l} 47.3 & \\
        & Raptor       &  6 & 147.4 & +0.00   & 518 $\times$ 518 & 44.2 & -3.1 \\
        & \shortmethod &  6 & \phantom{0}75.9 & -71.5 & 518 $\times$ 518 & 42.9 & -4.4 \\
        \midrule
        \SetCell[r=3]{c} \rotatebox{90}{\scriptsize DINOv3}
        & Baseline     & 12 & \SetCell[c=2]{l} 107.0 & & 512 $\times$ 512 &
        \SetCell[c=2]{l} 54.9 & \\
        & Raptor       &  6 & 107.0 & +0.00   & 512 $\times$ 512 & 49.1 & -5.8 \\
        & \shortmethod &  6 & \phantom{0}53.6 & -53.4 & 512 $\times$ 512 & 47.4 & -7.5 \\
        \bottomrule
    \end{tblr}
\end{table*}

\section{Related Work}

\paragraph{Depth as Recurrent Dynamics.} 
Existing work conceptualises deep networks as continuous dynamical systems where representations naturally cluster into meta-stable attractors as they propagate through depth \cite{veit2016residual, geshkovski2023the, karagodin2024clustering}. 
BRH exploits this simplicity bias by approximating contiguous blocks with a single parameter-tied layer applied recurrently \cite{jacobs2026block}, operating on the assumption that multi-step iterative execution is strictly necessary.
In contrast, we argue that this multi-step behaviour is merely an artefact of the optimisation landscape.
Instead of unrolling a recurrent operator, \shortmethod suggests that these redundant blocks approximate a single, ideal phase of computation that maps inputs directly to the terminal representation in a fused step.

\paragraph{Layer Relevance and Structural Pruning.} 
To handle block redundancy, current methods often identify ``ineffective'' layers and excise them using structural pruning \cite{gromov2025the, sajjad2023on, fan2019reducing}, accuracy-grounded relevance metrics \cite{hinostroza2026rethinking}, or structural linearisation of consecutive blocks \cite{ashkboos2024slicegpt, ma2023llmpruner, shopkhoev2025replaceme} to salvage downstream performance. 
Rather than simply dropping layers based on heuristics or generic proxies, we frame this multi-step processing explicitly as executable redundancy through the lens of \shortmethod. 
This theoretically justifies replacing an entire contiguous sequence of layers with a single step instead of selectively pruning isolated parts.

\paragraph{Trajectory Distillation and Block Collapse.} 
Advanced model compression techniques move beyond simple logit matching by distilling internal hidden-state trajectories (using metrics like CKA or value-relations) \cite{dasgupta2025improving, wang2020minilm} or by iteratively substituting full layers with retrained compact blocks \cite{musaeus2025iterative}. 
Building upon this, we introduce ``block collapse'' as a localised distillation mechanism guided by \shortmethod. 
By anchoring the start and end of a phase through deep supervision, we replace the entire redundant block with a single, non-recurrent surrogate operator. 
This forces the network to bypass intermediate representational dawdling, actively denoising the manifold and directly yielding inference-time compute savings.

\section{Conclusion}
\label{sec:conclusion}

In this work, we investigated the phenomenon of block redundancy in Vision Transformers.
While recent frameworks like BRH interpret highly similar contiguous layers as evidence of underlying recurrence, our analyses demonstrate that these iterative micro-steps are not strictly computationally necessary.
Our intrinsic dimension and variance analyses further suggest that, in the histopathology setting, intermediate steps within redundant phases can accumulate augmentation-sensitive variation without altering the coarse representational trajectory.

To resolve this inefficiency, we introduced \method (\shortmethod).
By dynamically identifying redundant phases and collapsing them into single, learned surrogate layers, \shortmethod replaces intermediate iterations with a single macro-step.
Unlike linear approximations that sacrifice token-mixing expressivity, or block-recurrent methods that still execute costly iterations at inference, our approach tracks the coarse geometric trajectory while reducing both parameter count and inference compute.
Across natural-image models, \shortmethod preserves competitive downstream performance at substantially reduced compute, while in several histopathology settings it matches or improves over the uncompressed baseline.

\subsection{Limitations and Further Work}
\label{subsec:limitations}

While \shortmethod provides a general methodology for exploiting block redundancy in ViTs, it is not to be interpreted as a universal panacea. 
As discussed in \Cref{tab:mil_results_h0_mini,tab:mil_results_hibou_b,tab:natimgcls}, performance improvement over the uncompressed baseline appears only in specific settings, notably the histopathology models in our experiments.
This does not imply that a performance increase should be expected in the general case.
Dense prediction remains more sensitive to block collapse than instance-level classification.
Our segmentation results suggest that patch-level tasks benefit from preserving more of the intermediate token trajectory, and may therefore require less aggressive collapse or task-aware distillation.

The partitioning scheme in \Cref{subsec:phase_discovery} remains an approximation. Better block discovery improves performance, motivating future work on submodular optimisation or optimal transport for this combinatorial problem.

Finally, we focus on post-hoc exploitation of block redundancy. Explaining how such redundancy arises in vision models, and how it might be mitigated during pre-training, remain important directions for future work.

% While the partitioning scheme outlined in \Cref{subsec:phase_discovery} is effective for discovering ideal phases, it is still an approximation. 
% Our results indicate that better block discovery choices have a positive effect, and while this problem is combinatorial, effective submodular optimisation strategies or optimal transport theory could be an interesting extension to further improve efficient exploitation of block redundancy.

% Finally, we emphasise that our focus in this work is on post-hoc exploitation of block redundancy, and a general framework for understanding computational phase discovery in vision models.
% This implies that several important questions fall somewhat out of scope: understanding how and why block redundancy occurs in vision models, and finding remedial strategies in pre-training would be an important direction for further work.

\subsubsection*{Acknowledgments}
This work was funded by the Research Council of Norway through Visual Intelligence, Centre for Research-based Innovation (309439), and by the South-Eastern Norway Regional Health Authority (2024039).
The computations were performed on resources provided by Sigma2 (NN8104K) --- the National Infrastructure for High-Performance Computing and Data Storage in Norway.
We acknowledge Sigma2 for access to the LUMI supercomputer, owned by the EuroHPC Joint Undertaking, hosted by CSC (Finland) and the LUMI consortium through Sigma2, Norway.

\bibliography{abrv,cv-ml-master,references}

% %%%%%%%%%%%%%%%%%%%%%%%%%%%%%%%%%%%%%%%%%%%%%%%%%%%%%%%%%%%%

\clearpage
\appendix
\renewcommand\thetable{\Alph{section}.\arabic{table}}
\renewcommand\thefigure{\Alph{section}.\arabic{figure}}
\counterwithin{figure}{section}
\counterwithin{table}{section}
\counterwithin{equation}{section}

\section{Histopathology Experimental Details}
\label{supp:histopathology}

\paragraph{Datasets and splits.}
For prostate cancer grading, we train MIL heads on PANDA~\cite{bulten2022artificial} and use the train/validation/test split from Song et al.~\cite{song2024morphological}. We use the validation split for checkpoint selection and report final performance on the PANDA test split. We use TCGA-PRAD~\cite{zuley2016radiology} only as an external evaluation cohort. For breast metastasis detection, we train MIL heads on CAMELYON17~\cite{litjens20181399} and evaluate on the CAMELYON17 test split and CAMELYON16~\cite{bejnordi2017diagnostic}. We use corrected CAMELYON slide labels from Ling et al.~\cite{ling2025comprehensive}.

\paragraph{Label processing.}
PANDA provides slide-level ISUP grade labels. For TCGA-PRAD, we convert Gleason primary and secondary patterns to ISUP grade groups using the standard mapping: Gleason $3+3$ maps to grade group 1, $3+4$ to 2, $4+3$ to 3, total score 8 to 4, and total score 9--10 to 5. We assign benign or non-cancer cases to grade group 0 when applicable. CAMELYON17 and CAMELYON16 use binary slide-level labels for metastasis detection.

\paragraph{Tiling and tissue filtering.}
We tile WSIs into non-overlapping $256\times256$ patches. For PANDA and TCGA-PRAD, we retain tiles with at least $60\%$ foreground tissue. For CAMELYON17 and CAMELYON16, we identify tissue regions using Otsu thresholding~\cite{otsu1975threshold}. During feature extraction, we resize tiles to the input resolution expected by each foundation model and apply the corresponding model-specific normalisation statistics.

\paragraph{Feature extraction.}
We freeze each pruned or unpruned backbone before MIL training and precompute tile embeddings for every downstream cohort. For DINOv2-style backbones with register tokens, we concatenate the class token with the mean of the patch tokens and exclude the four register tokens. This produces a $1536$-dimensional tile embedding. Each WSI bag contains all retained foreground tile embeddings.

\paragraph{MIL models.}
We train ABMIL~\cite{ilse2018attention} and TransMIL~\cite{shao2021transmil} on frozen tile embeddings with an identity encoder. For prostate grading, each MIL head predicts six ISUP classes. For metastasis detection, each MIL head predicts two classes. We use class-weighted cross-entropy for all MIL experiments.

\paragraph{Optimisation and checkpoint selection.}
For PANDA, we train each MIL head for 20 epochs using Adam with learning rate $10^{-4}$, weight decay $10^{-4}$, cosine annealing, batch size of one WSI, gradient clipping at norm 1, and gradient accumulation over 32 steps. We select the checkpoint with the highest PANDA validation QWK and evaluate it on PANDA test and TCGA-PRAD.

For CAMELYON17, we train each MIL head for 100 epochs using Adam with learning rate $10^{-4}$, weight decay $10^{-4}$, cosine annealing, class-weighted cross-entropy, and gradient clipping at norm 1. We evaluate after every epoch on CAMELYON17 validation, CAMELYON17 test, and CAMELYON16. We select the checkpoint with the highest 5-epoch moving average of CAMELYON17 validation accuracy and report the corresponding CAMELYON17 and CAMELYON16 accuracies.

\paragraph{Metrics.}
For PANDA and TCGA-PRAD, we report quadratic weighted kappa (QWK) multiplied by 100. For CAMELYON17 and CAMELYON16, we report slide-level accuracy in percentage points. We report the mean and standard deviation across independent MIL training runs.

\section{Histopathology Pruning Protocol}
\label{supp:pruning_protocol}

\paragraph{Phase discovery and pruning data.}
For the histopathology experiments, we compute phase boundaries using PANDA training tiles only. We sample $10{,}000$ tiles and compute layer-wise cosine distances between intermediate representations. We use the resulting phase boundaries to define the pruned \shortmethod students. All pruned backbones, including \shortmethod and the baselines, are trained by distillation on PANDA training tiles with the corresponding unpruned backbone frozen as the teacher. We do not use TCGA-PRAD, CAMELYON17, or CAMELYON16 for phase discovery or backbone pruning.

\paragraph{Optimisation.}
We train pruned backbones with feature-level MSE distillation, AdamW, cosine learning-rate decay, and gradient clipping. Unless stated otherwise, we use a learning rate $3\times10^{-4}$. NOSE, WDPruning, and \shortmethod use single-stage distillation. Raptor uses its standard two-stage protocol: recurrent block training followed by stitched model fine-tuning.

% \paragraph{Training budgets.}
% We choose the pruning training budget based on the size of the pruned feature extractor, using longer training for smaller students. For H0-mini, we train NOSE and WDPruning for 2 epochs, Raptor for 1 Stage-1 epoch plus 2 Stage-2 epochs, and \shortmethod for 2 epochs at depth 4 and 1.5 epochs at depth 5. For Hibou-B, we train NOSE and WDPruning for 5 epochs, Raptor for 3 epochs at depths 3 and 4, and \shortmethod for 2 epochs at depths 3 and 4.

\paragraph{Parameter counts.}
Parameter counts in the main tables refer to the feature extractor used at inference. For WDPruning, this includes only the retained first $K$ layers and the single depth-specific probe used for feature extraction. For NOSE, this uses the unbloated checkpoint after removing pruned attention parameters. For Raptor, this uses the recurrent feature extractor executed at inference.

\section{Ablation on Surrogate Initialisation}
\label{supp:ablation}
In the main text, we describe an auditioning process to select the optimal layer from the teacher's block to initialise the student's surrogate layer. To isolate the effect of this initialisation, \Cref{tab:worstbest} ablates the procedure by comparing models initialised with the best candidate versus the worst candidate within the block. The results show that the auditioned candidate generally provides stronger downstream transfer than the worst candidate, although the effect is not uniform across every dataset and compression depth, supporting the value of the auditioning step.

\begin{table*}[htb]
    \scriptsize
    \caption{Comparison of TWT initialised with the best vs.\ worst candidate block across all evaluated depths. We report QWK ($\times 100$) for prostate grading and accuracy for breast metastasis detection. Deltas compare against the baseline model.}
    \label{tab:worstbest}
    \centering
    \adjustbox{max width=\textwidth, keepaspectratio}{%
    \begin{tblr}{
      colspec = {c l l c *{4}{lr}}, 
      column{5-12} = {mode=math},
      column{6,8,10,12} = {font=\tiny, leftsep=-3pt},
      row{1,2} = {mode=text},
      row{4,6,9,11,14,16,19,21} = {bg=highlight},
    }
    \toprule
    \SetCell[c=4]{c} & & & & \SetCell[c=4]{c} Prostate (QWK) & & & & \SetCell[c=4]{c} Breast/Lymph (Acc) & & & \\
    \cmidrule[r]{5-8} \cmidrule[r]{9-12}
    Model & Agg. & Method & Layers & \SetCell[c=2]{l}{PANDA Test} & & \SetCell[c=2]{l}{TCGA-PRAD} & & \SetCell[c=2]{l}{CAM17 Test} & & \SetCell[c=2]{l}{CAM16} & \\
    \midrule
    \SetCell[r=10]{c} \rotatebox{90}{H0-mini} & \SetCell[r=5]{c} \rotatebox{90}{ABMIL} 
    & Baseline & 12 & \SetCell[c=2]{l} 93.77 \pm 0.16 & & \SetCell[c=2]{l} 69.14 \pm 1.22 & & \SetCell[c=2]{l} 88.35 \pm 0.72 & & \SetCell[c=2]{l} 97.31 \pm 0.23 & \\
    & & \shortmethod Best & 5 & 93.79 \pm 0.25 & \textcolor{teal}{+0.02} & 70.08 \pm 1.08 & \textcolor{teal}{+0.94} & 90.46 \pm 0.43 & \textcolor{teal}{+2.12} & 96.73 \pm 0.39 & \textcolor{red}{-0.57} \\
    & & \shortmethod Worst & 5 & 93.32 \pm 0.15 & \textcolor{red}{-0.45} & 68.61 \pm 0.33 & \textcolor{red}{-0.53} & 88.73 \pm 1.75 & \textcolor{teal}{+0.38} & 96.27 \pm 0.57 & \textcolor{red}{-1.04} \\
    & & \shortmethod Best & 4 & 93.71 \pm 0.28 & \textcolor{red}{-0.06} & 69.69 \pm 0.55 & \textcolor{teal}{+0.55} & 89.70 \pm 1.75 & \textcolor{teal}{+1.36} & 95.49 \pm 0.70 & \textcolor{red}{-1.81} \\
    & & \shortmethod Worst & 4 & 93.78 \pm 0.37 & \textcolor{teal}{+0.01} & 70.05 \pm 0.31 & \textcolor{teal}{+0.91} & 90.34 \pm 0.86 & \textcolor{teal}{+1.99} & 94.35 \pm 0.81 & \textcolor{red}{-2.95} \\
    \cmidrule{2-12}
    & \SetCell[r=5]{c} \rotatebox{90}{TransMIL} 
    & Baseline & 12 & \SetCell[c=2]{l} 94.57 \pm 0.41 & & \SetCell[c=2]{l} 56.02 \pm 6.21 & & \SetCell[c=2]{l} 89.11 \pm 0.71 & & \SetCell[c=2]{l} 96.27 \pm 0.98 & \\
    & & \shortmethod Best & 5 & 94.49 \pm 0.67 & \textcolor{red}{-0.08} & 61.89 \pm 3.56 & \textcolor{teal}{+5.87} & 90.13 \pm 0.71 & \textcolor{teal}{+1.02} & 95.29 \pm 0.67 & \textcolor{red}{-0.98} \\
    & & \shortmethod Worst & 5 & 94.76 \pm 0.32 & \textcolor{teal}{+0.19} & 58.93 \pm 8.55 & \textcolor{teal}{+2.91} & 89.19 \pm 0.95 & \textcolor{teal}{+0.08} & 96.43 \pm 0.96 & \textcolor{teal}{+0.16} \\
    & & \shortmethod Best & 4 & 94.29 \pm 0.59 & \textcolor{red}{-0.28} & 59.11 \pm 4.13 & \textcolor{teal}{+3.09} & 90.93 \pm 0.86 & \textcolor{teal}{+1.82} & 95.23 \pm 0.59 & \textcolor{red}{-1.04} \\
    & & \shortmethod Worst & 4 & 94.78 \pm 0.67 & \textcolor{teal}{+0.21} & 63.13 \pm 2.42 & \textcolor{teal}{+7.11} & 88.31 \pm 3.08 & \textcolor{red}{-0.80} & 93.52 \pm 1.70 & \textcolor{red}{-2.74} \\
    \midrule
    \SetCell[r=10]{c} \rotatebox{90}{Hibou-B} & \SetCell[r=5]{c} \rotatebox{90}{ABMIL} 
    & Baseline & 12 & \SetCell[c=2]{l} 93.70 \pm 0.24 & & \SetCell[c=2]{l} 66.27 \pm 1.11 & & \SetCell[c=2]{l} 90.25 \pm 0.87 & & \SetCell[c=2]{l} 90.52 \pm 2.48 & \\
    & & \shortmethod Best & 4 & 93.27 \pm 0.32 & \textcolor{red}{-0.44} & 67.77 \pm 0.67 & \textcolor{teal}{+1.50} & 90.25 \pm 1.03 & \textcolor{teal}{+0.00} & 96.16 \pm 0.28 & \textcolor{teal}{+5.64} \\
    & & \shortmethod Worst & 4 & 92.72 \pm 0.33 & \textcolor{red}{-0.99} & 66.78 \pm 0.48 & \textcolor{teal}{+0.50} & 89.49 \pm 1.35 & \textcolor{red}{-0.76} & 93.83 \pm 0.92 & \textcolor{teal}{+3.31} \\
    & & \shortmethod Best & 3 & 93.30 \pm 0.21 & \textcolor{red}{-0.41} & 65.49 \pm 0.60 & \textcolor{red}{-0.79} & 89.66 \pm 0.92 & \textcolor{red}{-0.59} & 93.63 \pm 1.43 & \textcolor{teal}{+3.11} \\
    & & \shortmethod Worst & 3 & 92.73 \pm 0.24 & \textcolor{red}{-0.98} & 64.14 \pm 1.13 & \textcolor{red}{-2.13} & 88.47 \pm 1.15 & \textcolor{red}{-1.78} & 93.83 \pm 0.74 & \textcolor{teal}{+3.31} \\
    \cmidrule{2-12}
    & \SetCell[r=5]{c} \rotatebox{90}{TransMIL} 
    & Baseline & 12 & \SetCell[c=2]{l} 94.13 \pm 0.44 & & \SetCell[c=2]{l} 57.32 \pm 7.11 & & \SetCell[c=2]{l} 89.32 \pm 0.65 & & \SetCell[c=2]{l} 92.80 \pm 2.99 & \\
    & & \shortmethod Best & 4 & 94.65 \pm 0.29 & \textcolor{teal}{+0.52} & 55.47 \pm 2.70 & \textcolor{red}{-1.85} & 91.02 \pm 1.12 & \textcolor{teal}{+1.70} & 93.26 \pm 2.36 & \textcolor{teal}{+0.46} \\
    & & \shortmethod Worst & 4 & 93.54 \pm 0.41 & \textcolor{red}{-0.58} & 59.23 \pm 4.82 & \textcolor{teal}{+1.91} & 90.00 \pm 1.09 & \textcolor{teal}{+0.68} & 90.93 \pm 2.11 & \textcolor{red}{-1.87} \\
    & & \shortmethod Best & 3 & 94.02 \pm 0.07 & \textcolor{red}{-0.11} & 54.58 \pm 3.12 & \textcolor{red}{-2.74} & 90.42 \pm 0.98 & \textcolor{teal}{+1.10} & 91.09 \pm 2.84 & \textcolor{red}{-1.71} \\
    & & \shortmethod Worst & 3 & 93.33 \pm 0.64 & \textcolor{red}{-0.80} & 49.70 \pm 7.65 & \textcolor{red}{-7.61} & 89.28 \pm 2.38 & \textcolor{red}{-0.04} & 91.29 \pm 0.95 & \textcolor{red}{-1.51} \\
    \bottomrule
    \end{tblr}
    }%
\end{table*}

\FloatBarrier

\section{Sensitivity of Discovery Threshold $\varepsilon$}
\label{supp:sensitivity}
\begin{figure}[htb]
\centering
\includegraphics[width=0.8\linewidth]{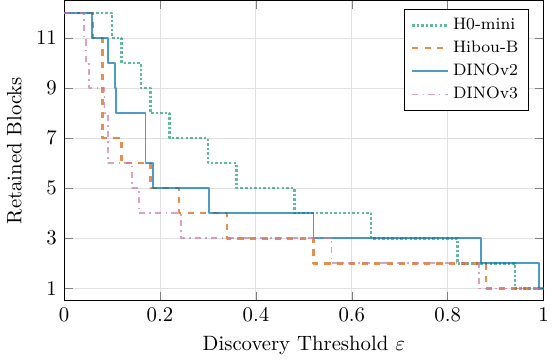}
\vspace{5pt}
\caption{Sensitivity analysis showing the effect of the discovery threshold $\varepsilon$ on the total number of retained blocks after partitioning.}
\label{fig:sensitivity}
\end{figure}

The threshold $\varepsilon$ serves as the core hyperparameter for dynamic block discovery. Rather than a direct training parameter, it acts as a geometric distance bound: smaller $\varepsilon$ enforces strict boundary requirements, yielding highly granular partitions with many retained blocks, whereas larger values lead to more collapse. \Cref{fig:sensitivity} illustrates this monotonic relationship across our four evaluated backbones, showing how varying $\varepsilon$ affects the count of retained blocks.

\end{document}